\documentclass[sigconf]{acmart}
\usepackage[ruled]{algorithm2e} 
\usepackage{algorithm2e}
\usepackage{amsmath, amsfonts}  
\usepackage{enumerate}
\usepackage{enumitem}
\usepackage{graphicx}
\usepackage{hyperref}
\usepackage{wrapfig}
\usepackage{epstopdf}
\usepackage{my-macro}
\usepackage{url}
\usepackage{listings}
\theoremstyle{definition}

\usepackage{amsthm}
\usepackage{multirow}
\usepackage{subfigure}

\newcommand{\ours}{{\sf STEAM}\xspace}

\def \b {\mathbf{b}}
\def \d {\mathbf{d}}

\def \u {\mathbf{u}}
\def \W {\mathbf{W}}

\def \D {\mathcal{D}}

\AtBeginDocument{%
  \providecommand\BibTeX{{%
    \normalfont B\kern-0.5em{\scshape i\kern-0.25em b}\kern-0.8em\TeX}}}

\setcopyright{acmcopyright}
\copyrightyear{2018}
\acmYear{2018}
\acmDOI{10.1145/1122445.1122456}

\copyrightyear{2020}
\acmYear{2020}
\setcopyright{acmcopyright}\acmConference[KDD '20]{Proceedings of the 26th ACM SIGKDD Conference on Knowledge Discovery and Data Mining}{August 23--27, 2020}{Virtual Event, CA, USA}
\acmBooktitle{Proceedings of the 26th ACM SIGKDD Conference on Knowledge Discovery and Data Mining (KDD '20), August 23--27, 2020, Virtual Event, CA, USA}
\acmPrice{15.00}
\acmDOI{10.1145/3394486.3403145}
\acmISBN{978-1-4503-7998-4/20/08}

\begin{document}



\title{STEAM: Self-Supervised Taxonomy Expansion with Mini-Paths}


\author{Yue Yu}
\affiliation{%
  \institution{Georgia Institute of Technology}
  \streetaddress{P.O. Box 1212}
  \city{Atlanta}
  \state{GA}
  \country{USA}
  \postcode{30332}
}
\email{yueyu@gatech.edu}

\author{Yinghao Li}
\affiliation{%
  \institution{Georgia Institute of Technology}
  \streetaddress{P.O. Box 1212}
  \city{Atlanta}
  \state{GA}
  \country{USA}
  \postcode{30332}
}
\email{yinghaoli@gatech.edu}

\author{Jiaming Shen}
\affiliation{%
  \institution{University of Illinois at Urbana-Champaign}
  \city{Urbana}
  \state{IL}
  \country{USA}
  }
\email{js2@illinois.edu}

\author{Hao Feng}
\affiliation{%
  \institution{University of Electronic Science and Technology of China}
  \city{Chengdu}
  \state{Sichuan}
  \country{China}
  }
\email{is.fenghao@gmail.com}

\author{Jimeng Sun}
\affiliation{%
  \institution{University of Illinois at Urbana-Champaign}
  \city{Urbana}
  \state{IL}
  \country{USA}
  }
\email{jimeng@illinois.edu}

\author{Chao Zhang}
\affiliation{%
  \institution{Georgia Institute of Technology}
  \city{Atlanta}
  \state{GA}
  \country{USA}
  \postcode{30332}
}
\email{chaozhang@gatech.edu}




\begin{abstract}





Taxonomies are important knowledge ontologies that underpin numerous
applications on a daily basis, but many taxonomies used in practice suffer from
the low coverage issue. We study the taxonomy expansion problem, which aims to
expand existing taxonomies with new concept terms. We propose a self-supervised
taxonomy expansion model named \ours, which leverages natural supervision in the
existing taxonomy for expansion. To generate natural self-supervision signals,
\ours samples mini-paths from the existing taxonomy, and formulates a node
attachment prediction task between anchor mini-paths and query terms. To solve
the node attachment task, it learns feature representations for query-anchor
pairs from multiple views and performs multi-view co-training for prediction.
Extensive experiments show that \ours outperforms state-of-the-art methods for
taxonomy expansion by $11.6\%$ in accuracy and $7.0\%$ in mean reciprocal rank on three public benchmarks. The implementation of \ours can be found at \url{https://github.com/yueyu1030/STEAM}.




\end{abstract}

\newif\ifsubmit
\submitfalse

\ifsubmit
\newcommand{\chao}[1]{}
\newcommand{\yue}[1]{}
\newcommand{\yinghao}[1]{}
\newcommand{\js}[1]{}
\newcommand{\jiaming}[1]{}

\else
\newcommand{\zc}[1]{{\color{red}[Chao: #1]}}
\newcommand{\yue}[1]{{\color{teal}[Yue: #1]}}
\newcommand{\yinghao}[1]{{\color{cyan}[Yinghao: #1]}}
\newcommand{\js}[1]{{\color{purple}[JS: #1]}}
\newcommand{\jiaming}[1]{{\color{green}[Jiaming: #1]}}
\fi

\newcommand{\insertt}{$\uparrow$}
\newcommand{\substitute}{$\rightarrow$}
\newcommand{\commentt}{\textasciitilde}
\newcommand{\deletee}{$\downarrow$}

\newcommand{\temp}{{\sf \textsc{STEAM}}\xspace}
\newcommand{\ourco}{{\sf \textsc{STEAM-Co}}\xspace}
\newcommand{\ourdep}{{\sf \textsc{STEAM-C}}\xspace}
\newcommand{\ouremb}{{\sf \textsc{STEAM-D}}\xspace}
\newcommand{\ourpat}{{\sf \textsc{STEAM-L}}\xspace}

\newcommand{\taxo}{\mathcal{T}}
\newcommand{\termSet}{\mathcal{V}}
\newcommand{\edgeSet}{\mathcal{E}}
\newcommand{\corpus}{\mathcal{D}}
\newcommand{\newTermSet}{\mathcal{C}}
\newcommand{\newEdgeSet}{\mathcal{R}}
\newcommand{\query}{q}
\newcommand{\anchorTerm}{p}
\newcommand{\miniPath}{P}
\newcommand{\pathLength}{L}
\newcommand{\pathSet}{\mathcal{P}}
\newcommand{\positiveSet}{\mathcal{X}^{\rm pos}}
\newcommand{\positivePair}{X^{\rm pos}}
\newcommand{\negativeSet}{\mathcal{X}^{\rm neg}}
\newcommand{\trainingSet}{\mathcal{X}}
\newcommand{\trainingPair}{X}
\newcommand{\childTerm}{a}
\newcommand{\childSet}{A}
\newcommand{\negativeTerm}{n}
\newcommand{\negativeTermSet}{\mathcal{N}}
\newcommand{\positionn}{y}

\setlength{\abovedisplayskip}{3pt plus 1pt minus 1pt}
\setlength{\belowdisplayskip}{3pt plus 1pt minus 1pt}



\maketitle

\section{Introduction}


Concept taxonomies play a central role in a
wide spectrum of applications.  On a daily basis, e-commerce websites like
Amazon heavily rely on their product taxonomies to support billions of product
navigations, searches, and recommendations \cite{zhang2014taxonomy}; scientific taxonomies (\eg, MeSH\footnote{https://www.nlm.nih.gov/mesh/meshhome.html}) make it much faster to
identify relevant information from massive scientific papers, and concept
taxonomies in knowledge bases (\eg, Freebase \cite{bollacker2008freebase}) underpin many question answering systems
\cite{harabagiu2003open}.
Due to such importance, many taxonomies have been curated in general and specific domains, \eg,
WordNet \cite{miller1995wordnet}, Wikidata \cite{vrandevcic2012wikidata}, MeSH~\cite{lipscomb2000medical}, Amazon Product Taxonomy~\cite{karamanolakis2020txtract}.

One bottleneck of many existing taxonomies is the {\it low coverage problem}. This problem arises mainly due to two reasons.  First, many existing taxonomies are curated by domain experts. As the curation process is expensive and time-consuming, the result taxonomies often include only frequent and coarse-grained terms. Consequently, the curated taxonomies have high precision, but limited coverage. Second, domain-specific knowledge is constantly growing in most applications.
New concepts arise continuously, but it is too tedious to rely on human
curation to maintain and update the existing taxonomies. The low
coverage issue can largely hurt the performance of downstream tasks, and
automated taxonomy expansion methods are in urgent need.

\begin{figure*}[t]
  \centering

  \includegraphics[width=0.95\textwidth]{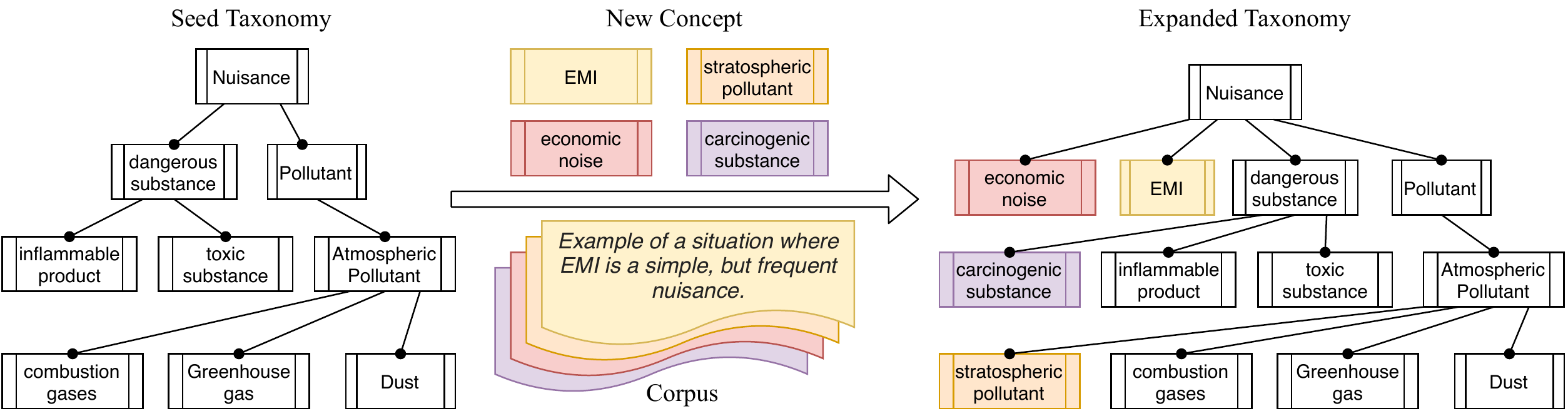}
  \caption{Illustration of the taxonomy expansion problem. Given an existing
    taxonomy, the task is to insert new concept terms (\eg, \textit{EMI}, \textit{stratospheric pollutant}, \textit{economic noise}, \textit{carcinogenic substance}) into the correct positions in the existing taxonomy.
  }
  \label{fig:temp_overall}

\end{figure*}


Existing taxonomy construction methods follow two lines. One line is to
construct taxonomies in an unsupervised way
\cite{zhang2018taxogen,panchenko2016taxi,liu2012automatic, wang2013phrase}. This
is achieved by hierarchical clustering \cite{zhang2018taxogen}, hierarchical topic
modeling \cite{liu2012automatic,wang2013phrase}, or syntactic patterns (\eg, the Hearst
pattern~\cite{hearst1992automatic}). The other line adopts supervised approaches
\cite{mao2018end,gupta2017taxonomy, kozareva2010semi}, which first detect
hypernymy pairs (\ie, term pairs with the ``\textit{is-a}'' relation) and then
organize these pairs into a tree structure. However, applying these methods for
taxonomy \emph{expansion} suffers from two limitations. First, most of them
attempt to construct taxonomies \emph{from scratch}. Their output taxonomies can
rarely preserve the initial taxonomy structures curated by domain experts.
Second, the performance of many methods relies on large amounts of annotated
hypernymy pairs, which can be expensive to obtain in practice.

We propose a self-supervised taxonomy expansion model named \ours\footnote{Short
for \textbf{S}elf-supervised \textbf{T}axonomy \textbf{E}xp\textbf{A}nsion with
\textbf{M}ini-Paths.}, which leverages natural supervision in the existing
taxonomy for expansion. To generate natural self-supervision signals, \ours
samples \emph{mini-paths} from the existing taxonomy, and formulates a node
attachment prediction task between mini-paths and query terms.
The
 \emph{mini-paths}, which contain terms
in different layers (\eg ``\emph{Pollutant}''--``\emph{Atmospheric
Pollutant}''--``\emph{Dust}'' in Figure \ref{fig:temp_overall}), serve as
candidate \textit{anchors} for query terms and yield
many training \textit{query-anchor} pairs from the existing taxonomy. With these query-anchor pairs, we learn a model  (Section
\ref{sect:minipath}) to
pinpoint the correct position for a query term in the mini-path.
Compared with previous
methods~\cite{shwartz2016improving,vedula2018enriching,shen2020taxoexpan}
using  single anchor terms, \ours
better leverages the existing taxonomy since the mini-paths contain richer
structural information from different levels.

In cooperation with mini-path-based node attachment, \ours extracts features for
query-anchor pairs from multiple views, including: (1) \emph{distributed features} that capture the similarity between terms' distributed representations; (2) \emph{contextual features}, \ie information from two terms' co-occurring sentences; (3) \emph{lexico-syntactic features} extracted from the similarity of surface string names between terms.
We find that different views can provide
complementary information that is vital to taxonomy expansion.
To fuse the three views more effectively, we propose a multi-view
co-training procedure  (Section \ref{sect:cotrain}).  In this procedure, the three views lead to different
branches for predicting the positions of the query term, and the predictions
from these three views are encouraged to agree with each other.



We have conducted extensive experiments on three taxonomy construction benchmarks in different domains. The results show that \ours outperforms
state-of-the-art methods for taxonomy expansion by $11.6\%$ in accuracy and
$7.0\%$ in mean reciprocal rank.  Moreover, ablation studies demonstrate the effect of mini-path for capturing structural information from the taxonomy, as well as the multi-view co-training for harnessing the complementary signals from all views.

Our main contributions are: 1) a self-supervised framework that performs
taxonomy expansion with natural supervision signals from existing taxonomies
and text corpora; 2) a mini-path-based anchor format that better captures
structural information in taxonomies for expansion; 3) a multi-view co-training
procedure that integrates multiple sources of information in an end-to-end
model; and 4) extensive experiments on several benchmarks verifying the efficacy of our method.

\section{Problem Description}



We focus on the taxonomy expansion task for \emph{term-level} taxonomies, which is formally defined as follows.

\begin{definition}[Taxonomy] { A taxonomy $\taxo = (\termSet, \edgeSet)$ is a tree structure
 where 1) $\termSet$ is a set of terms (words or phrases); and 2)
 $\edgeSet$ is a set of edges representing \emph{is-a} relations between
 terms. Each directed edge $\langle v_i, v_j\rangle \in \edgeSet$ represents a hypernymy
 relation between term $v_i$ and term $v_j$, where $v_i$ is the hyponym (child)
 and $v_j$ is the hypernym (parent).}
 \end{definition}



 The problem of taxonomy expansion (Figure \ref{fig:temp_overall}) is to enrich an initial taxonomy by inserting
new terms into it. These new terms are often automatically extracted and
filtered from a text corpus. Formally, we define the problem as below:

\begin{definition}[Taxonomy Expansion]  Given 1)  an existing taxonomy
$\taxo_{0} = (\termSet_0, \edgeSet_0)$, 2) a text corpus
$\corpus$, and 3) a set of candidate terms $\newTermSet$, the goal of
taxonomy expansion is to insert the term $\query\in \newTermSet$ into the existing
taxonomy $\taxo_{0}$ and expand it into a more complete
taxonomy $\taxo = (\termSet, \edgeSet )$ where $\termSet = \termSet_{0}\cup \newTermSet, \edgeSet =\edgeSet_0 \cup \newEdgeSet$
with $\newEdgeSet$ being the newly discovered relations between terms in
$\newTermSet$ and $\termSet_0$.  \end{definition}

\section{The \ours Method}

In this section, we describe our proposed method \ours. We first give an
overview of our method, and then detail the two key modules: mini-path-based
prediction and multi-view co-training. Finally, we discuss the model learning
and inference procedures.


\subsection{Self-Supervised Learning by Mini-Path Attachment}
\label{sect:minipath}


The central task of taxonomy expansion is to attach a query term $\query \in \newTermSet$ into the
correct position in the existing taxonomy $\taxo_0$. \ours learns to attach query terms using \emph{natural} supervision signals from the seed taxonomy. Its self-supervised learning procedure aims to preserve the structure of the seed taxonomy by creating a learning task that pinpoints the anchor positions for the terms already seen in the seed taxonomy. The training data for this self-supervised learning task can be easily obtained from the seed taxonomy, thereby facilitating learning a model that performs query attaching at inference time.





\subsubsection{Query-Anchor Matching with Mini-Paths}



To instantiate the self-supervised learning paradigm~\cite{liang2020beyond,shen2020taxoexpan,tung2017self}, an intuitive idea is to
find the best hypernym for the query term $\query$.  Most existing works
~\cite{shen2020taxoexpan,vedula2018enriching,mao2018end} follow this idea and
model the taxonomy expansion problem as finding the optimal hypernym pairs for
test terms. They usually design a binary classifier trained by determining whether
 $\langle \anchorTerm_i, \anchorTerm_j \rangle$ ($\anchorTerm_i, \anchorTerm_j
 \in \termSet_0$) is a hypernymy pair.


Unlike the binary classification formulation, \ours learns to match query terms
with anchors with richer structural information.  The core of \ours's
self-supervised learning procedure is \emph{mini-paths}, which are snippets
sampled from the seed taxonomy.  These mini-paths, containing the term pairs
from different layers of taxonomy, can preserve the hierarchical relations
among different terms.  Below, we introduce the notion of \emph{mini-path} and
formulate the self-supervised learning task based on mini-paths.

\begin{definition}[Mini-path]
  A mini-path $\miniPath = [\anchorTerm_1, \anchorTerm_2, \ldots, \anchorTerm_{L} ]$ consists of several terms $\{\anchorTerm_1, \anchorTerm_2, \ldots, \anchorTerm_{L} \} \subset \termSet_{0}$, where $\pathLength$ is the length of $\miniPath$. Each term pair $\langle \anchorTerm_i, \anchorTerm_{i+1}\rangle \ (1\leq i \leq \pathLength-1)$ corresponds to an edge in  $\edgeSet_{0}$.
\end{definition}




\begin{figure}[h]
    \centering {\
        \subfigure[An illustration of mini-paths.] {
            \label{subfig:temp_train_a}
            \includegraphics[width=0.45\textwidth]{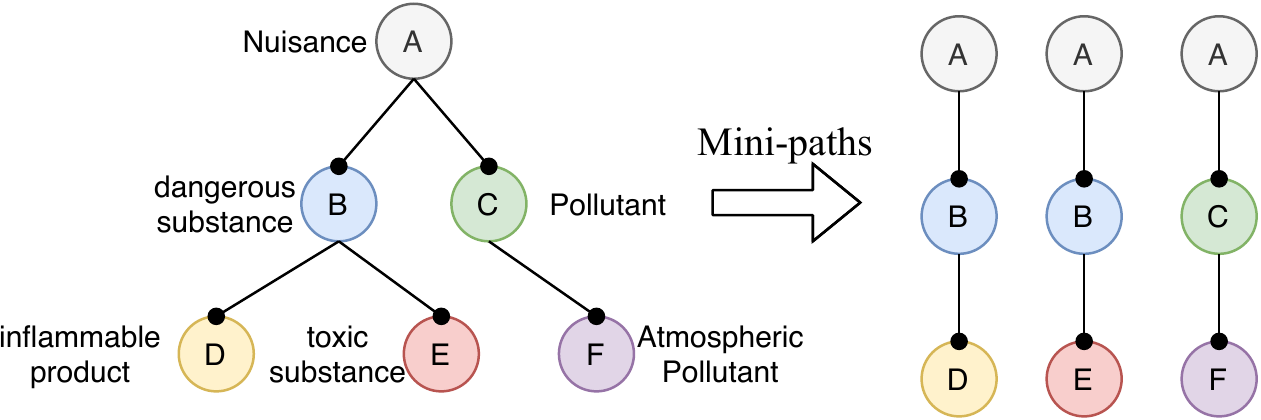}
        }
        \subfigure[The classification target.] {
            \label{subfig:temp_train_b}
            \includegraphics[width=0.45\textwidth]{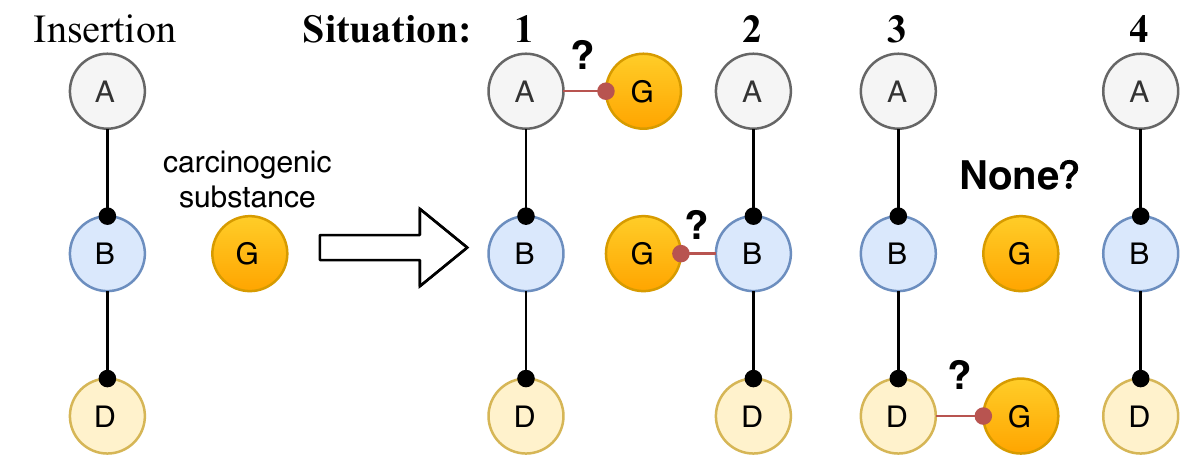}
        }
    }
            
    \caption{An illustration of the proposed self-supervised data structure, including the construction of mini-paths and the learning target during the term-insertion process.}
    \vspace{-3mm}
    \label{fig:temp_train}
    
\end{figure}

The mini-paths are fixed-length branchless sub-graphs of the existing taxonomy $\taxo_0$, as shown in Figure~\ref{subfig:temp_train_a}, which maintain part of the parent-child relationships between terms.
To keep the hierarchical information in the taxonomy with the self-supervised training
set, we design the training task as a multi-class classification problem.
As shown in Figure \ref{subfig:temp_train_b}, given a 3-terms mini-path as \emph{anchor} and a new term as \emph{query}, \ours predicts the probabilities of the \emph{query} being attached to the three terms or none of them.

Compared with the simple task of binary hypernymy classification, matching
query terms with mini-paths has two major advantages: 1) When attaching a query
term $\query$ into an anchor mini-path $\miniPath$, we consider the collection of all terms $\anchorTerm_i\in\miniPath$ as a whole, rather than attend to them separately.
This not only provides richer information for query attachment but also
results in larger training data for self-supervised learning.
2) Compared with
the binary classification, this task is more challenging---the matching module
needs to judge not only whether $\query$ should be matched to $\miniPath$ but also which specific position to attach.
Learning from this more challenging self-supervised task allows \ours to better leverage the structural
information of the existing taxonomy and perform better for anchor term deduction and taxonomy expansion.



\subsubsection{Sampling Mini-Paths from the Seed Taxonomy}


To facilitate the learning problem, we need to sample mini-paths from the existing
taxonomies as anchors, as well as the query terms that should be attached
to different positions in anchor mini-paths. This can be achieved by randomly
sampling mini-paths in the seed taxonomy, along with positive and
negative query terms for each mini-path.


The detailed procedure for training data creation is described as follows.
Given one mini-path $\miniPath \in \pathSet$ where $\pathSet$ is the collection of all mini-paths in the existing taxonomy,
we first generate positive training set $\positiveSet$ by sampling all the child terms
$\childTerm_{i,l}\in\childSet$  of  $\miniPath \in \pathSet$, where $\childTerm_{i,l}$ is the $i$-${\rm th}$ child of the $l$-${\rm th}$ anchor term $\anchorTerm_l\in\miniPath$ and $\childSet$ contains all child terms attached to the mini-path, and a positive pair is represented as $\positivePair_{i,l} = \langle \childTerm_{i,l}, \anchorTerm_j, l \rangle$.
Once $\positiveSet$ is obtained, we augment the training set by adopting the negative sampling strategy to generate negative set $\negativeSet$ by randomly selecting $|\negativeSet| = r\times|\positiveSet|$ terms $\{\negativeTerm_i\}_{i=1}^{|\negativeSet|} $ with sampling ratio $r$, each constituting a negative pair with one term  that is not its parent in a anchor.
Since these negative terms do not directly associate with the mini-path $\miniPath$, we assign a relative position $\pathLength+1$ for them to indicate no connection exists.
Combining $\positiveSet$ and $\negativeSet$ together we obtain the final training set $\trainingSet$.




After obtaining  query-anchor pairs, we
need to learn a model using such data.  Given the set of training
pairs $\trainingSet$, we denote each pair as $\trainingPair = \langle \query, \miniPath, l
\rangle \in \trainingSet$ where $\query$ is the query term, $\miniPath$ is the
mini-path, and $\positionn$ is the relative position and aim to learn a
model $f(\query, \miniPath|\Theta)$ to identify the correct position
(represented by the true label $\positionn$).
The training objective is to optimize the negative log likelihood
$
    \ell = -\sum_\trainingPair \sum_{i=1}^{\pathLength+1} \positionn_i \log \hat{\positionn_i}
$
where $\hat{\positionn}$ is the predicted position.


\subsection{Multi-View Co-Training with Mini-Paths}
\label{sect:cotrain}


Now the question is how to obtain feature representations for each query-anchor
pair $(q, P)$. In \ours, we learn feature representations for query-anchor
pairs from three different views and integrate them with a multi-view
co-training procedure.


\subsubsection{Multi-View Feature Extraction}

\ours learns representations of query-anchor pairs from three views: (1)
\emph{the distributed representation view}, which captures their correlation
from pre-trained word embeddings; (2) \emph{the contextual relation view}, which
captures their correlation from the sentences where the query term and anchor
terms co-occur; and (3) \emph{the lexico-syntactic view}, which captures their
correlation from the linguistic similarities between the query and the anchor.


Each of the three views has its own advantages and disadvantages: (1)
\emph{Distributed features} have a high coverage over the term vocabulary, but
they do not explicitly model pair-wise relations between a query term and an
anchor term; (2) \emph{Contextual features} can capture the relation between two
terms from their co-occurred sentences, but have limited coverage over term
pairs. For example, only less than 15\% of hypernym pairs have co-occurred in
the scientific corpus of the SemEval dataset; (3) \emph{Lexico-Syntactic
features} encode linguistic information between terms and can work well for
matched term pairs, but these features are too rigid to cover all the linguistic
patterns, and may also have limited coverage.

Given a query term $q$ and an anchor mini-path $P = [p_1, p_2, \cdots, p_L]$,
we describe the details of how we learn representations for the query-anchor
pair $(q, P)$ from the three different views.



\para{(1) Distributed Features.}  
The first view extracts distributed features
for both the query $q$ and the anchor mini-path $P$.  For the query term $q$ and the anchor terms in the mini-path $P$, we
use pre-trained BERT embeddings~\cite{devlin2019bert} to initialize their
distributed representations due to its superior expressive power~\cite{jiang2019smart,liang2020bond}. While it is feasible to directly use such initial
embeddings for similarity computation, recent work \cite{shen2020taxoexpan}
shows that the neighboring terms of an anchor term are also useful for taxonomy
expansion. We follow \cite{shen2020taxoexpan} to use a position-enhanced graph attention
network (PGAT) to propagate the embeddings for the terms in the seed taxonomy by
considering the taxonomy as a directed graph---this
will lead to updated embeddings for the anchor terms in the mini-path $P$.  For each anchor term $p_l \in P$, we use $\w(p_l)$ to denote
its PGAT-propagated embedding and use $\w(q)$ to denote the embedding of
the query term $q$, then we concatenate these embeddings and obtain the
distributed representation for the query-anchor pair $(q, P)$: 
\begin{equation}
  \h_{d}(q, P) = [ \w(q) \oplus \w(p_1) \oplus \cdots \oplus \w(p_L)].
\label{eq1}
\end{equation}






\para{(2) Contextual Features.} When two terms co-occur in the same sentence, the contexts of their co-occurrence can often indicate the relation of the two terms. Our second view thus harvests the sentences from the given corpus $D$ to extract features for the query term $q$ and the mini-path $P$.
Given the query term $q$ and any anchor term $p_l \in P$, we fetch all the sentences where $q$ and $p_l$ have co-occurred from corpus $D$. Similar to
\cite{shwartz2016improving}, we process these sentences to extract the
dependency paths between $q$ and $p_l$ in these sentences, denoted as $\D_{q,
p_l}$.  For each dependency path ${d}_{q, p_l} \in \D_{q, p_l}$, it is a
sequence of context words that lead $q$ to $p_l$ in the dependency tree:
\begin{equation}
  d_{q, p_l} = \{v_{e_1}, v_{e_2}, \cdots, v_{e_k} \},
\end{equation}
where $k$ is the length of the dependency path. 
Each edge $v_e$ in the
dependency path contains 1) the connecting term $v_l$, 2) the part-of-speech tag of
the connecting term $v_{pos}$, 3) the dependency label $v_{dep}$, and 4) the edge
direction between two subsequent terms $v_{dir}$. Formally, each edge $v_e$ is
represented as: $v_e = [v_l, v_{pos}, v_{dep}, v_{dir}]$.
Now in order to encode each extracted dependency path $d_{q, p_l}$, we feed the
multi-variate sequence  $d_{q, p_l}$ into an LSTM encoder. The representation
of the LSTM's last hidden layer, denoted as $\text{LSTM}(d_{q, p_l})$, is then used as the
representation the path $d_{q, p_l}$. As the set $\D(q, p_l)$ contains multiple
dependency paths between $q$ and $p_l$, we aggregate them with the attention
mechanism to compute the weighted average of these path representations:
\begin{equation}
\begin{aligned}
  \hat{\alpha}_{d}& = \u^{T} \tanh \left(\W \cdot \text{LSTM}(d_{q, p_l})\right), \\
  \alpha_d &=\frac{\exp \left( \hat\alpha_d\right)}{\sum_{d' \in \D_{q, p_l}} \exp \left(\hat\alpha_{d'}\right)},\\
  \d(q, p_l)&= \sum_{d \in \D(q, p_l)} \alpha_{d} \cdot \text{LSTM}(d_{q, p_l}),\\
\end{aligned}
\end{equation}
where $\alpha_d$ denotes attention weight for the dependency path $d_{q, p_l}$; $\W, \u$ are
trainable weights for the attention network.

The final
contextual features between $q$ and $P$ is thus given by
\begin{equation}
  \h_{c}(q, P) = [ \d(q, p_1) \oplus  \cdots  \oplus \d(q, p_L)].
\label{eq2}
\end{equation}



\para{(3) Lexical-Syntactic Features.} Our third view extracts
lexical-syntactic features between terms. Such features encode the correlations
between terms based on their surface string names and syntactic information
\cite{mao2018end,panchenko2016taxi,zhang2016learning}. 
Given a term pair $(x,y)$, we extract the following lexical-syntactic features
between them:
\begin{itemize}[leftmargin=*]
    \item \textbf{Ends with}:  Identifies whether $y$ ends with $x$ or not. 
    \item \textbf{Contains}: Identifies whether $y$ contains $x$ or not. 
    \item \textbf{Suffix match}: Identifies whether the $k$-length suffixes of $x$ and $y$ match or not. 
    \item \textbf{LCS}: The length of longest common substring of term $x$ and $y$.
    \item \textbf{Length Difference}: The normalized length difference between $x$ and $y$. Let the length of term $x$ and $y$ be $L(x)$ and $L(y)$, then the normalized length difference is calculated as $\frac{|L(x)-L(y)|}{\max \left(L(x), L(y)\right)}$.
    \item \textbf{Normalized Frequency Difference}: The normalized frequency of $(x, y)$ in corpus $D$ with min-max normalization. Specifically, follow~\cite{gupta2017taxonomy}, we consider \textbf{two types of} normalized frequency based on the noisy hypernym pairs obtained in ~\cite{panchenko2016taxi}: (1) \emph{the normalized frequency difference}. Given a term pair $(x,y)$, their normalized frequency is defined as $nf(x,y) = \frac{freq(x, y)}{\max_{z \in \mathcal{V}}freq(x,z)}$ where $freq(x,y)$ defines the occurrence frequency between term $(x,y)$ in the hypernym pairs given by \cite{panchenko2016taxi} and $\mathcal{V} = \mathcal{V}_0 \cup \mathcal{C}$ which is all terms in the existing taxonomy and test set. Then the first normalize frequence difference is defined as $f(x,y) = nf(x,y)-nf(y,x)$. (2) \emph{the generality difference}. For term $x$, the normalized generality score $ng(x)= log(1+h)$, where $h$
is defined as the logarithm of the number of its distinct hyponyms. Then the generality difference of  term pair $g(x,y)$ is defined as the difference in generality between $(x,y)$ as $g(x,y)=ng(x) - ng(y)$.
\end{itemize}

Given the query term $q$ and the mini-path $P = [p_1, p_2, \cdots, p_L]$, we
compute the lexico-syntactic features for each pair $(q,p_l)~(1 \le l \le
L)$, denoted as $\s(q, p_l)$ and concatenate the features derived from all
the term pairs as the lexical-syntactic features for  $(q, P)$:
\begin{equation}
  \h_{s}(q, P) = [ \s(q, p_1) \oplus  \cdots  \oplus \s(q, p_L)].
\label{eq3}
\end{equation}

 \begin{figure*}
     \centering
     \includegraphics[width=0.95\textwidth]{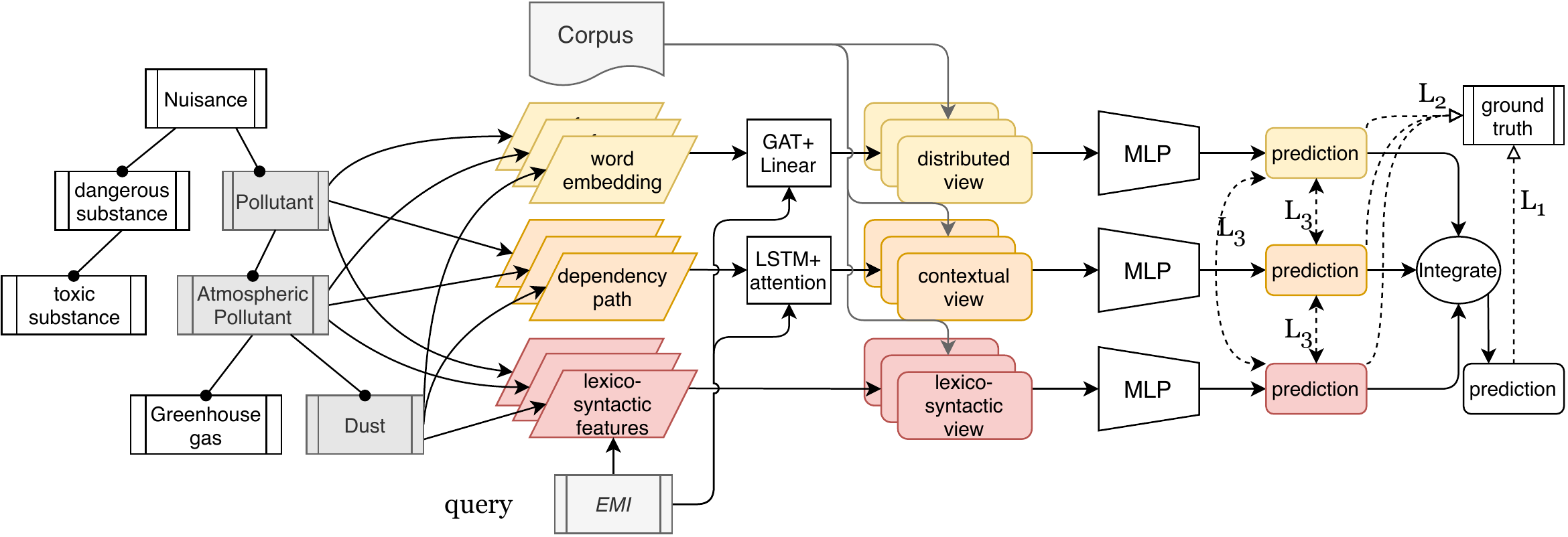}
     \caption{Illustration of the proposed co-training model architecture.
     The grey terms in the existing taxonomy on the left is an anchor path to attach the new term to.
     $L_1$, $L_2$ and $L_3$ corresponds to the log-likelihood loss and Euclidean loss calculated in Equation \eqref{eq:l1}, \eqref{eq:l2} and \eqref{eq:l3} respectively. 
     }
    \vspace{-2mm}
     \label{fig:temp}
 \end{figure*}

\subsubsection{The Multi-View Co-Training Objective}

As the three views provide complementary information to each other, it is
important to aggregate the three views for the query-anchor matching.
To this end, one may simply stack three different sets of features and train one unified classifier~\cite{mao2018end}. However, such  feature-level integration can lead to suboptimal results due to two reasons: (1) one view can provide dominant signals over the other two, making it hard to fully unleash the discriminative power of each view;  (2) the three views can have different dimensionality and distributions, making learning a unified classifier from concatenated features difficult.

To more effectively harvest the information from the three different views, we propose a multi-view co-training procedure. This co-training procedure (see Figure \ref{fig:temp}) uses the three views to learn three different classifiers and then derives an aggregated classifier from the three classifiers and also encourages their predictions to be consistent. 
The entire model can be trained in an end-to-end manner. Below, we first describe the base classifiers designed for the three different views, and then present the co-training objective.


\para{Base Classifiers from Multiple Views.} Based on three sets of feature
$\h_{d}, \h_{c}, \h_{s}$ derived from different views, we design three neural
classifiers for the query-anchor matching task, \ie, the multi-class
classification problem formulated in Section 
\ref{sect:minipath}. 
For each of the three views, we use a multi-layer perceptron (MLP) with one hidden layer
for this prediction task, denoted as $f_d$, $f_s$, and $f_r$.
Then the predictions from the three views are given by:\\
\begin{equation}
\begin{aligned}
    \y^d &= f_d(\h_{d}) = \W_2^d (\sigma(\W_1^d\h_{d}+\b_1^d) + \b_2^d ), \\
    \y^c &= f_c(\h_{c}) = \W_2^c (\sigma(\W_1^c\h_{c}+\b_1^c) + \b_2^c ),\\
    \y^s &= f_c(\h_{s}) = \W_2^s (\sigma(\W_1^s\h_{s}+\b_1^s) + \b_2^s ),
    \\
\end{aligned}
    \label{eq:4}
\end{equation}
where $\{{\W}_{1}^{k}, {\W}_{2}^k, \b_1^k, {\b}_{2}^k \}$ $k \in \{d,s,c\}$ are
trainable parameters for the three MLP classifiers, and $\sigma(\cdot)$ is the
activation function for which we use ReLU in our experiments.

\para{The Co-Training Objective.} Figure~\ref{fig:temp} shows the co-training
model that integrates the three base classifiers.  From the three base
classifiers $f_d$, $f_s$, and $f_r$, we design an aggregated
classifier for the final output. This aggregated classifier, which we denote as
$f_\text{agg}$, integrates the three base classifiers by averaging over their
predictions\footnote{We have also tried to use attention mechanism to aggregate the score but didn't see an obvious performance gain.}:
\begin{equation}
  \y^\text{agg} = f_\text{agg} \left(\y^d, \y^c, \y^s \right) 
  =  \text{softmax} \left(\frac{1}{3} \left(\y^d + \y^s + \y^r \right) \right).
\label{ref:agg}
\end{equation}



To jointly optimize the base classifiers as well as the aggregated classifier, we develop a co-training procedure that not only learns the classifiers to fit the self-supervised signals but also promotes consistency among these classifiers.  The co-training objective involves three types of supervision, as detailed below.

The first loss $\ell_1$ is defines for the aggregated
classifier $f^\text{agg}$, which produces the final output.  Let $\{(\x_i, {\y}_i\}_{i=1}^N$ be
  the training dataset, where $\x_i$ is a query-anchor pair and ${\y}_i$ is
  the label indicating the correct position of the query term in the anchor
  mini-path.  Then  $\ell_1$ is defined as the negative log-likelihood loss:
\begin{equation}
\label{eq:l1}
  \ell_{1}=-\sum_{i=1}^N \sum_{j=1}^C {\y}_{ij} \log \y_{ij}^\text{agg},
\end{equation}
where $C=L+1$ is the number of labels for query-anchor matching.


The second loss $\ell_2$ is defined for three base classifiers corresponding to
the three views:
	\begin{equation}
	\label{eq:l2}
          \ell_{2}=-\sum_{u \in \{d,c,s\}} \sum_{i=1}^N \sum_{j=1}^C {\y}_{ij} \log \y_{ij}^u.
	\end{equation}

  The third loss $\ell_3$  is a
\emph{ consistency loss } that  encourages the prediction results from different views to
agree with each other. We use L2-distance to measure the difference between the classifiers and define $\ell_3$  as:
\begin{equation}
\label{eq:l3}
  \ell_{3}= \sum_{u, v \in \{d,s,r\}} \sum_{i=1}^N \left\|{\y}_i^u-{\y}_i^v\right\|^{2}.
\end{equation}

The overall objective of our model is then:
\begin{equation}
  \ell = \ell_1 + \lambda \ell_2 + \mu \ell_3,
  \label{eq:14}
\end{equation}
where $\lambda>0, \mu>0$ are two pre-defined balancing hyper-parameters.

\subsection{Model Learning and Inference}
During training, we learn the model parameter $\Theta$  by
minimizing the total
loss $\ell$ using stochastic gradient optimizers such as Adam \cite{kingma2014adam}.
During inference, given a new query term $\query \in \newTermSet$, we
traverse all the mini-paths $\miniPath \in \pathSet$ and calculate the
scores for all anchor terms $\anchorTerm \in \miniPath$ based on the aggregated final
prediction score $\positionn_{q, p}^{P}$ in Eq.~\eqref{ref:agg}.
Specifically, for any anchor term $\hat{p}$, we calculate its score of being the parent of query $q$ as
\begin{equation}
    y_{\hat{p}}= \frac{1}{|\hat{\mathcal{P}}|}\sum_{P \in \hat{\mathcal{P}}}{y_{q, \hat{p}}^{P}},
\end{equation}
where $\hat{\mathcal{P}}$ is the set of mini-paths which contain term $\hat{p}$.  Then, we
rank all anchor terms and select the term $p^*$ with the highest score as the predicted parent of the query $q$:
\begin{equation}
    p^* = \arg\max_{p \in \mathcal{V}_0} y_{p}.
\end{equation}

\subsection{Complexity Analysis}
At the training stage, our model uses $|\mathcal{P}|$ training instances every epoch and thus scales linearly to the number of mini-paths in the existing taxonomy. From above we have listed the number of mini-paths in our training, and the number of such mini-paths is linear to $O(|\mathcal{V}_0|)$ (\ie the number of terms in the existing taxonomy). At the inference stage, for each query term, we calculate $L|\mathcal{P}|$  matching scores, where $L$ is the length of the mini-path. To accelerate the computation, we use GPU for matrix multiplication and pre-calculate distributional and lexico-syntactic features and store the dependency paths for faster evaluation. 

\vspace{-1mm}
\section{Experiments}
\vspace{-1mm}
In this section, we evaluate the empirical performance of our proposed \ours
method. Our experiments are designed to answer the following three research
questions:
\begin{itemize}[leftmargin=*]
    \item \textbf{RQ1}: How well does \ours perform for the taxonomy expansion
      task compared with state-of-the-art methods?
    \item \textbf{RQ2}: How effective are the two key components in \ours:
      mini-path-based prediction and multi-view co-training?
    \item \textbf{RQ3}: What are the effects of different parameters on the performance of \ours?
\end{itemize}

\subsection{Experiment Setup}

\subsubsection{Datasets}

We evaluate the performance of our taxonomy construction method using three
public benchmarks.  These datasets come from the shared task of taxonomy
construction in SemEval 2016 \cite{bordea2016semeval}. We use all the three
English datasets in SemEval 2016, which correspond to three human-curated
concept taxonomies from different domains: environment (EN), science (SCI), and
food (Food). The statistics of these three benchmarks are presented in table~\ref{tb:dataset}. For each taxonomy, we start from the root term and randomly grow in a
top-down manner until 80\% terms are covered. We use the randomly-growed
taxonomies as seed taxonomies for self-supervised learning, and the rest 20\%
terms as our test data.
Our \ours method and several baselines also require external text corpora to
model the semantic relations between concept terms.

\begin{table}[t]
\vspace{-2mm}
\centering
\caption{The statistics of the three datasets for evaluation.}
\renewcommand\tabcolsep{3.2pt}
\begin{tabular}{cccc}
\toprule
Dataset      & Environment & Science & Food \\ \midrule
\# of Terms  & 261         & 429     & 1486 \\
\# of Edges & 261         & 452     & 1576 \\
\# of Layers & 6           & 8       & 8    \\ \bottomrule
\end{tabular}
\label{tb:dataset}
\vspace{-2mm}
\end{table}





\subsubsection{Baselines}
We compare with the following baselines: 
\begin{itemize}[leftmargin=*]

\item \textbf{TAXI}\footnote{\url{https://github.com/uhh-lt/taxi}} \cite{panchenko2016taxi} is a taxonomy induction method that reached the first place in the SemEval 2016 task. It first extracts hypernym pairs based on substrings and lexico-syntactic patterns with domain-specific corpora and then organizes these terms into a taxonomy. 
\item \textbf{HypeNet}\footnote{\url{https://github.com/vered1986/HypeNET}} \cite{shwartz2016improving} is a strong hypernym extraction method, which uses an LSTM-CNN model to jointly model the distributional and relational information between term pairs.
\item \textbf{BERT+MLP} is a distributional method for hypernym detection based  on pre-trained BERT embeddings. For each term pair, it first obtains term  embeddings from a pre-trained BERT model, and then feeds them into a Multi-layer Perceptron to predict whether they have the hypernymy relationship\footnote{For combining term embeddings, we experimented with \textsc{Concat}, \textsc{Difference}, and \textsc{Sum} as different fusing functions and report the best performance.}.
\item \textbf{TaxoExpan}\footnote{\url{https://github.com/mickeystroller/TaxoExpan}} \cite{shen2020taxoexpan} is state-of-the-art
  self-supervised method for taxonomy expansion. It adopts graph
  neural networks to encode the positional information and
  uses a linear layer to identify whether the candidate term is the parent of  the query term. For a fair comparison, we also use BERT embeddings for TaxoExpan instead of the word embeddings as in the original paper.
\end{itemize}




\subsubsection{Variants of ~{\ours}}
We also compare with several variants of \ours to
evaluate the effectiveness of its different modules:
\textsc{Concat} directly concatenates the three features and feeds it into an MLP for prediction; \textsc{Concat-D} concatenates only the context and lexico-syntactic views; \textsc{Concat-C} concatenates the distributed and the lexico-syntactic features; \textsc{Concat-L} concatenates the distributed and the context features; \textbf{\ourco} directly uses the aggregated classifier for prediction instead of the co-training objective (\ie, $\lambda=\mu=0$); \textbf{\ouremb} co-trains without the distributed view; \textbf{\ourdep} co-trains without the contextual view and \textbf{\ourpat} co-trains without the lexico-syntactic view.


\subsubsection{Implementation Details}

All the baseline methods, except for BERT-MLP, are obtained from the code
published by the original authors. The others (BERT-MLP, our model, and its
variants) are all implemented in PyTorch.
When learning our model, we use the ADAM optimizer ~\cite{kingma2014adam} with
a learning rate of 1e-3.  On all the three datasets, we train the model for 40
epochs as we observe the model has converged after 40 epochs. To prevent
overfitting, we used a dropout rate of 0.4 and a weight decay of 5e-4.  For
encoding context features, we follow \cite{shwartz2016improving} and set the
dimensions for the POS-tag vector, dependency label vector, and edge direction
vector, to 4, 5, and 1, respectively; and set the dimension for hidden units in
the LSTM encoder to 200. For the three base MLP classifiers, we set the
dimensions of the hidden layers to 50.  For sampling negative mini-paths, we set the size of negative samples $r=4$.
In the co-training module, there are
two key hyper-parameters: $\lambda$ and $\mu$ for controlling the strength for
training base classifiers and the consistency among classifiers. By default, we
set $\lambda=0.1, \mu=0.1$. We will study how these parameters affect the
performance of our model later.





\subsubsection{Evaluation Protocol} 
At test time, pinpointing the correct parent for a query term is a ranking problem.   
Specifically, given $n$ test samples, let us use $\{\hat{y}_1, \hat{y}_2, \cdots,
\hat{y}_n\}$ to denote their ground truth positions, $\{y_1, y_2, \cdots,
y_n\}$ to denote model predictions. Follow existing works~\cite{Manzoor2020expand,shen2020taxoexpan,vedula2018enriching}, we use multiple metrics as follows:\\
(1) \textbf{Accuracy (Acc)} measures the exact match accuracy for terms in the test set. It only counts the cases when the prediction equals to the ground truth, calculated as
    $$\text{Acc} = \frac{1}{n}\sum_{i=1}^{n}\mathbb{I}(y_i=\hat{y}_i).$$ \\
    (2) \textbf{Mean reciprocal rank (MRR)} is the average of reciprocal ranks of a query concept's true parent among all candidate terms.    Specifically, it is calculated as
    $$    \text{MRR} = \frac{1}{n}\sum_{i=1}^n \frac{1}{{rank}(y_{i})}.$$\\
    (3) \textbf{Wu \& Palmer similarity (Wu\&P)} calculates the semantic similarity between the predicted parent term $y$ and the ground truth parent term $\hat{y}$ as
    $$
        \omega\left( \hat{y}, y \right) = \frac{2\times \text{depth}(\text{LCA}(\hat{y}, y))}{\text{depth}(\hat{y}) +  \text{depth}(y)} 
    $$
    where ``$\text{depth}(\cdot)$'' is the depth of a term in  the taxonomy and ``$\text{LCA}(\cdot, \cdot)$'' is the least common ancestor of the input terms in the taxonomy. Then, the overall Wu\&P score is the mean Wu \& Palmer similarity for all terms in the test set written as $\text{Wu\&P} = \frac{1}{n}\sum_{i=1}^{n}\omega(y_i, \hat{y}_i)$.



\subsection{Experimental Results}

\subsubsection{Comparison with baselines.} 

\begin{small}
\begin{table}[t]
    \centering
    \caption{Comparision of \ours against the baseline methods on the three datasets (in \%). To reduce the randomness, we ran all methods for three times and report the average performance. TAXI outputs an entire taxonomy instead of ranking lists, so we are unable to obtain the MRRs for it.
    }
    \renewcommand\tabcolsep{2.2pt}
    \begin{tabular}{c| ccc| ccc| c c c}
    \toprule
    Dataset & \multicolumn{3}{c|}{\textbf{Environment}} &\multicolumn{3}{c|}{\textbf{Science}} & \multicolumn{3}{c}{\textbf{Food}}\\
    \midrule
     Metric & Acc & MRR & Wu\&P  & Acc & MRR  & Wu\&P &  Acc  & MRR  & Wu\&P \\
    \midrule
    BERT+MLP & 11.1 & 21.5 & 47.9 & 11.5 & 15.7 & 43.6& 10.5 & 14.9&47.0 \\
    TAXI &  16.7& -- & 44.7 & 13.0 & -- & 32.9 & 18.2 & -- & 39.2 \\
    HypeNet & 16.7 & 23.7 &  55.8 & 15.4 & 22.6 & 50.7  & 20.5 & 27.3& 63.2 \\
    TaxoExpan & 11.1 & 32.3 & 54.8  & 27.8  & 44.8 & 57.6 &  27.6 & 40.5 & 54.2 \\
    \ours & \textbf{36.1} &\textbf{46.9} & \textbf{69.6} & \textbf{36.5} & \textbf{48.3} & \textbf{68.2} &\textbf{34.2} & \textbf{43.4} & \textbf{67.0}\\
    \bottomrule 
    \end{tabular}
    \label{tb:full_guidance_results}
\end{table}
\end{small}

Table \ref{tb:full_guidance_results} reports the performance of \ours and the
baseline methods on the three benchmarks. From the results, we have the
following observations:

\noindent $\bullet$ \ours consistently outperforms all the baselines by large
margins on the three datasets. In particular, \ours improves the performance of
the state-of-the-art TaxoExpan model by 11.6\%, 7.0\% and 9.4\% for Acc, MRR and
Wu\&P on average. Such improvements are mainly due to the mini-path-based
prediction and the multi-view co-training designs in \ours.

\noindent $\bullet$ 
Among the baselines, TaxoExpan achieves the strongest overall performance.
The key advantage of TaxoExpan compared with other baselines is that it
propagates the embeddings among neighbors in the taxonomy via graph neural
networks. From the results, we can see that embedding propagation is 
effective in improving the MRR, making it achieve close MRRs with \ours.
However, TaxoExpan is largely outperformed by \ours in accuracy. This
phenomenon shows that while distributed features are useful for finding
relevant concepts,  contextual  and lexico-syntactic features are important
for pinpointing the exact hypernymy relationships.






\noindent $\bullet$ 
Pre-trained BERT embeddings have remarkable expressive power. However, BERT embeddings alone can yield limited
performance in the taxonomy expansion task since BERT does not well capture the contextual relations and between
terms.  \ours is based on BERT embedding, but it integrates contextual and
pattern information, which are highly useful for improving the performance.

\noindent $\bullet$
TAXI underperforms other methods on all three datasets. The major
drawback of TAXI and other taxonomy construction methods is that they
fail to use self-supervision signals in the  existing taxonomy. This
hinders them from learning the hierarchical and semantic information. Moreover, they simply use lexico-syntactic patterns and
neglect other distributional features, which is important for taxonomy
expansion.


\noindent $\bullet$ 
HypeNet outperforms BERT and TAXI since it combines the contextual and
distributed features. However, it neglects the structural information during
training and does not consider lexico-syntactic features, rendering it less
effective than \ours.

\subsubsection{Ablations Studies}
\begin{small}
\begin{table}[t]
    \centering
    \caption{Overall results of all variants of our methods on three datasets (in \%).
    }
    \renewcommand\tabcolsep{2.2pt}
    \begin{tabular}{c| ccc| ccc| c c c}
    \toprule
    Dataset & \multicolumn{3}{c|}{\textbf{Environment}} &\multicolumn{3}{c|}{\textbf{Science}} & \multicolumn{3}{c}{\textbf{Food}}\\
    \midrule
     Metric & Acc & MRR & Wu\&P  & Acc & MRR  & Wu\&P &  Acc  & MRR  & Wu\&P \\
    \midrule
    \textsc{Concat}& 25.0 & 40.3 & 64.2 & 20.4 & 25.8 &51.1 & 15.5 & 23.8 &49.6\\
    \textsc{Concat-D}& 30.6 & 38.6 & 63.7 & 11.1 & 20.1 & 48.1 & 23.1 & 28.9 &55.4\\
    \textsc{Concat-C}& 27.7 & 37.4 & 57.8 & 13.5 & 25.7 & 53.3 & 25.3 & 31.2 &58.3\\
    \textsc{Concat-L}& 11.1 & 31.4 & 57.7 & 13.5 & 23.7 & 39.1 & 8.30 & 13.4 &40.1\\
    \ourco & 25.0 & 41.0 & 66.3 & 32.7 & 45.3 & 64.4 & 31.1 & 40.7 & 65.1 \\
    \ouremb &  13.8& 32.0&	54.3 & 23.1&32.9&	60.0& 20.1 & 31.5 & 60.8 \\
    \ourdep & 11.1	&26.8&	49.2 &32.7&	44.5&	67.2 & 19.3 & 29.7 & 59.3\\
    \ourpat & 11.1&	27.5 &	51.6& 23.1&	36.5&	62.1 & 12.7 & 22.6 & 56.7\\
    \ours & \textbf{36.1} &\textbf{46.9} & \textbf{69.6} & \textbf{36.5} & \textbf{48.3} & \textbf{68.2} &\textbf{34.2} & \textbf{43.4} & \textbf{67.0}\\
    \bottomrule
    \end{tabular}
    \label{tb:full_guidance_result}

\end{table}
\end{small}
We perform ablation studies to study the effectiveness of the
different components in \ours: 1) mini-path-based self-supervised learning; 2)
the multi-view information; and 3) the co-training procedure.

\begin{figure}[t]
        \centering
        \subfigure[Environment]{
            \includegraphics[width=0.146\textwidth]{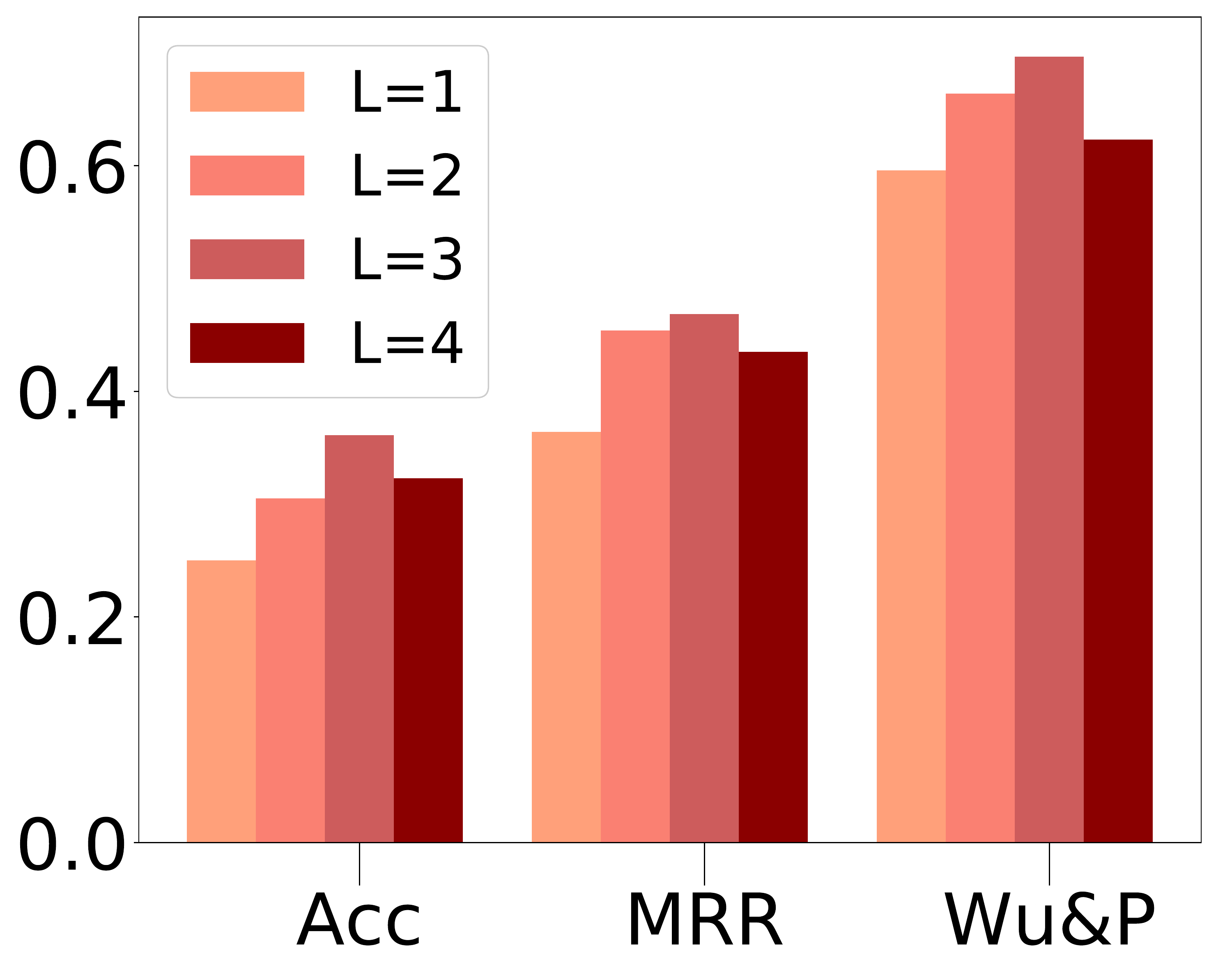}
            \label{fig:App_usage}
        }
        \subfigure[Science]{
            \includegraphics[width=0.146\textwidth]{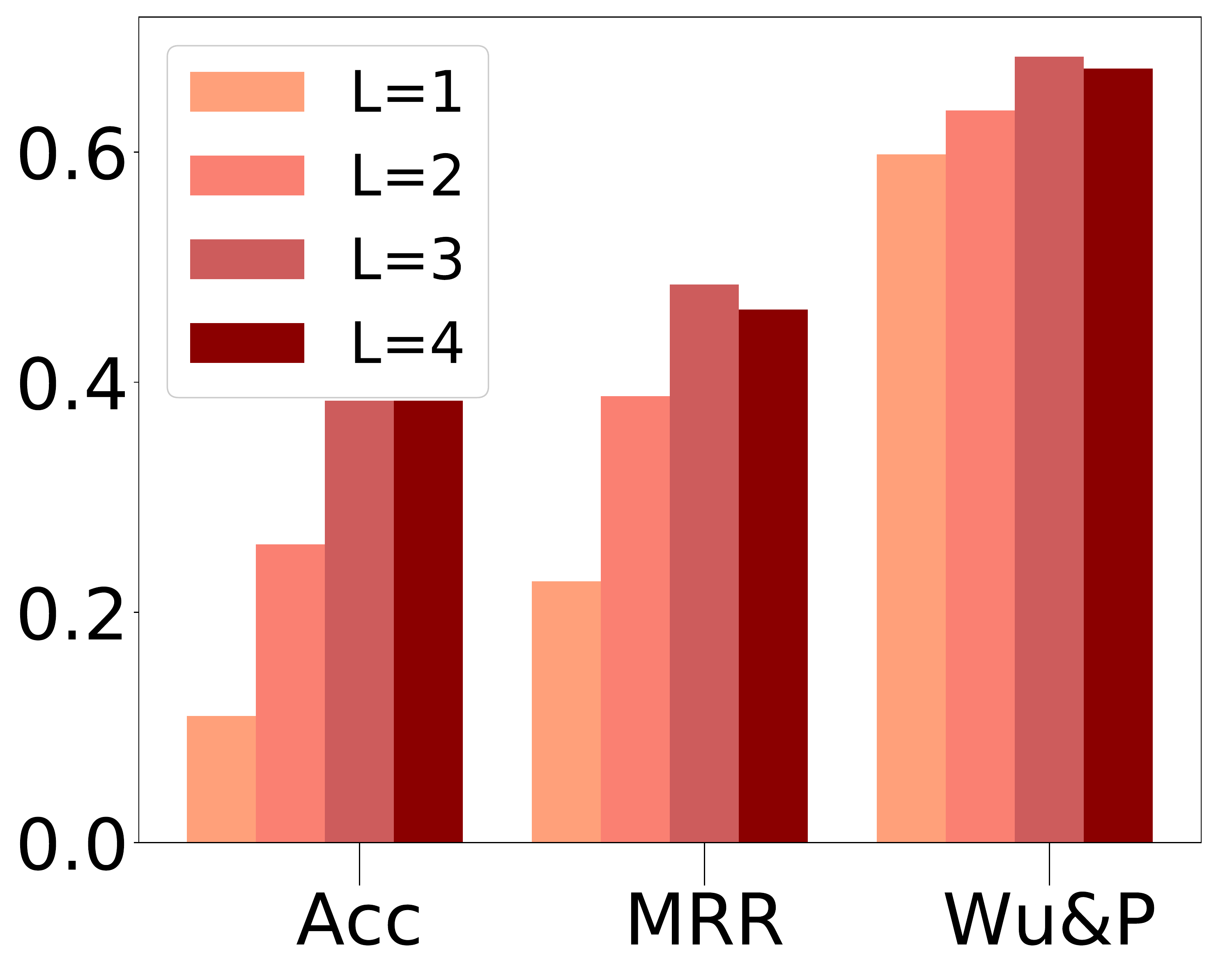}
            \label{fig:App_loc}
        }
        \subfigure[Food]{
            \includegraphics[width=0.146\textwidth]{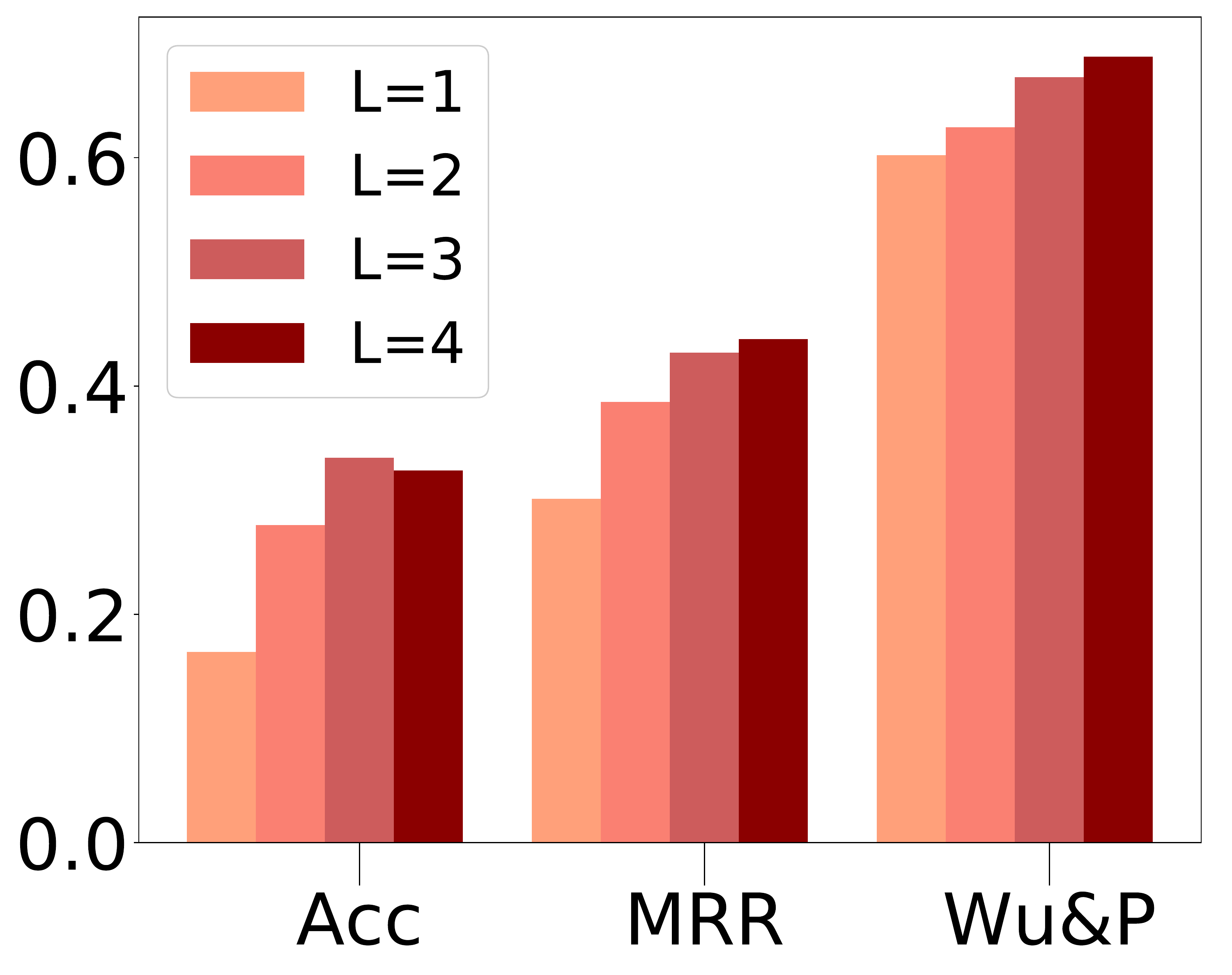}
            \label{fig:App_day}
        }

        \caption{The result for different length of mini-paths $L$ over three datasets.}\label{fig:data}

\end{figure}

\noindent \textbf{The Effect of Mini-Paths.} To study the effectiveness of
mini-path-based self-supervised expansion, we vary the length $L$ of mini-paths.
Note that, when $L=1$, the model is reduced to performing hypernymy prediction.
Figure \ref{fig:data} shows the performance of \ours on the three datasets when
$L$ varies. Generally, when $L$ is small, the performance of \ours stably
increases with $L$. Such results show that mini-paths can effectively capture
the structural information in the seed taxonomy---apart from the `parent' of the
query term, the grandparents and siblings contain additional information to
improve expansion performance. The mini-paths connect terms from different
layers of the taxonomy and carry such information to make the model pinpoint the
correct position. However, when $L$ increases from 3 to 4, we observe slight performance drops. This is because the size of the training data shrinks
for smaller taxonomies when $L$ becomes larger. Take the environment dataset as
an example: It contains 185 training samples when $L=3$ while 83 when $L=4$. As
a result, the final performance decreases by $3.2\%$ for accuracy.






\noindent \textbf{The Effect of Multi-view Information.} We study the
contributions of different views by comparing \ours with its variants (\ouremb,
\ourdep, \ourpat). Table \ref{tb:full_guidance_result} shows the results on the
three datasets. As shown, it is clear that all three types of features
contribute significantly to the overall performance. When eliminating one of the
three views, the average performance drops by 6.07\%, 8.10\% and 4.67\% for the
three metrics.




\noindent \textbf{The Effect of Co-training.} Now we proceed to study the
effectiveness of the co-training procedure. While integrating multiple views is
important, \emph{how} to integrate multi-view information is equally important.
From the results in Table \ref{tb:full_guidance_results}, one can see \ours
outperforms \textsc{Concat} by $15.3\%$, $16.2\%$ and $13.3\%$ for three
metrics on average.  This verifies the effectiveness of co-training
compared with concatenation: the simple concatenation strategy cannot fully
harvest the information from each view and could make the learning problem more
difficult.   Interestingly, the performance for \textsc{Concat} is even worse
than \textsc{Concat-D} and \textsc{Concat-C} in accuracy on Food and
Environment, which implies that simple concatenation can even hurt the
performance with more views.

The co-training objective in \ours involves two loss terms that encourage better learning of the base classifiers and the consistency among them. From
Table \ref{tb:full_guidance_results}, the performance gap between \ours and \ourco shows the effectiveness of these two terms. \ourco only uses the aggregated classifier for prediction and underperforms \ours by large margins.
The reason is that these terms explicitly require \emph{every} base classifier is sufficiently trained and mutually enhances each other; without them, certain views may not be fully leveraged, which limit the effectiveness in leveraging multi-view information for training.

\subsubsection{Parameter Studies}

In this subsection, we study the effects of different parameters on the
performance of \ours. We have already studied the effect of the path length in
the ablation study, now we study the effects of two key parameters in the
co-training procedure: 1) the weight of the prediction loss of the three base
classifiers $\lambda$, and 2) the weight of the consistency loss $\mu$. When
evaluating one parameter, we fix other parameters to their default values and
report the results. Due to the space limit, we only report the results on
parameters on Science dataset as the tends and findings are similar for the
three datasets.

\begin{figure}[t]
    \vspace{-2mm}
        \centering
        \subfigure[$\lambda$]{
    \includegraphics[width=0.222\textwidth]{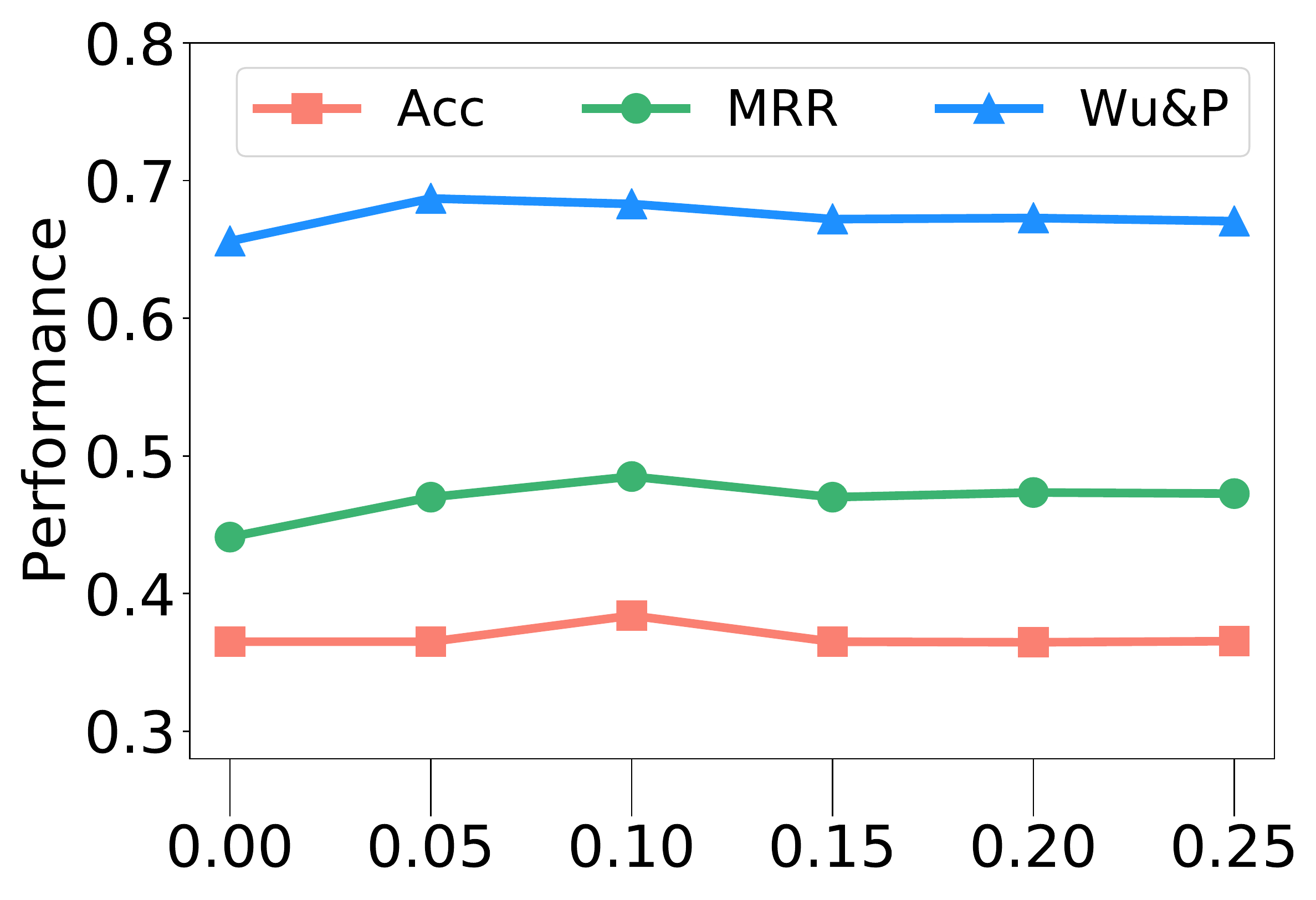}
            \label{fig:lambda}
        }
        \subfigure[$\mu$]{
            \includegraphics[width=0.222\textwidth]{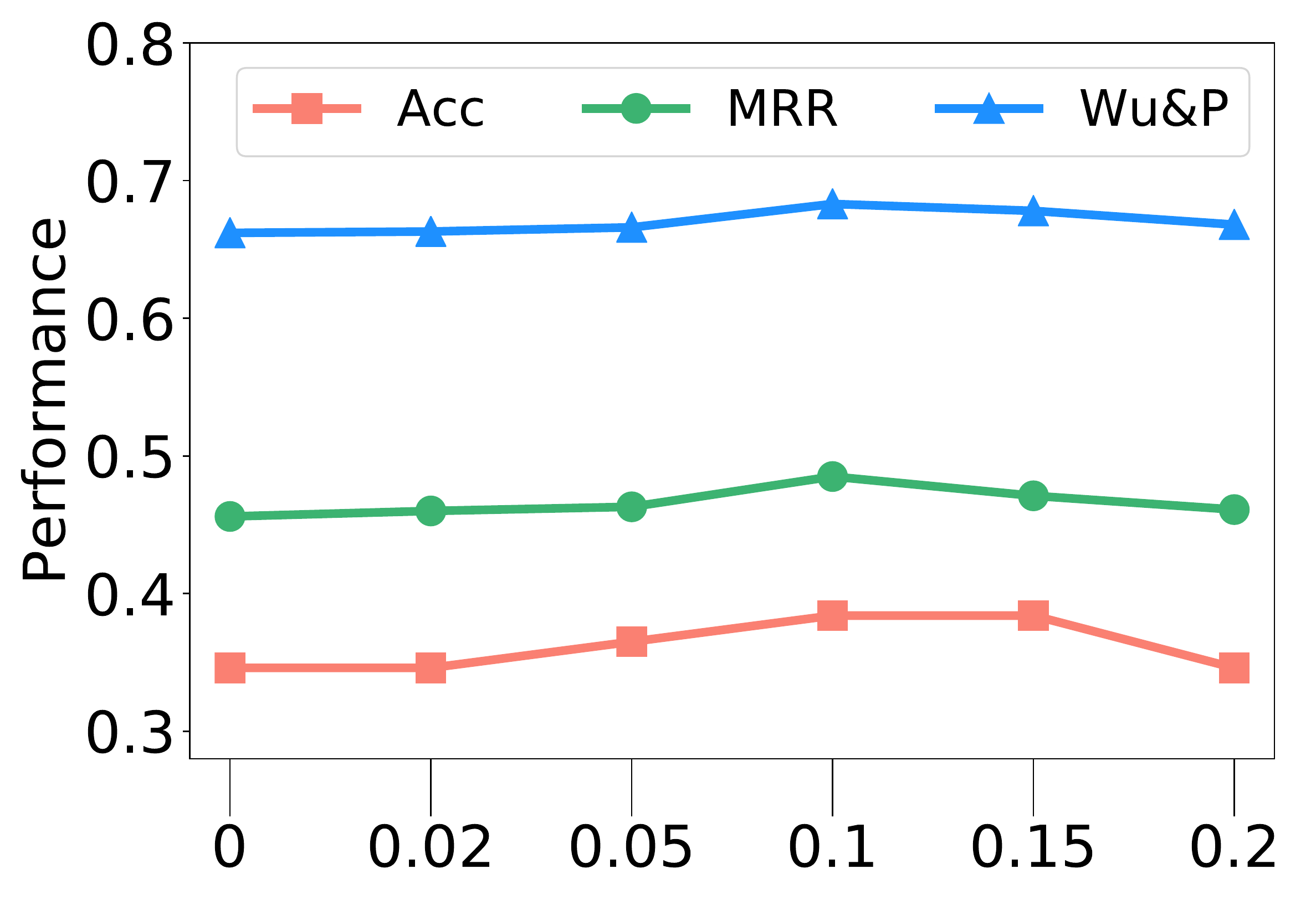}
            \label{fig:mu}
        }
            \vspace{-1ex}
        \caption{The performance of our model when varying different parameters.}\label{fig:param}
\end{figure}

\noindent \textbf{Effect of $\lambda$.} Figure \ref{fig:lambda} shows the
effect of $\lambda$ on the Science dataset. We can observe that as $\lambda$
increases, the performance improves for all three metrics. This is because
larger $\lambda$ will add more weight to learning base classifiers and enforce
each base classifier to achieve good prediction performance. As the base
classifiers become stronger, the derived aggregated classifier can also become
stronger. However, when $\lambda \geq 0.15$, the performance decreases with
$\lambda$. We suspect the reason is each single view can be one-sided and noisy
to yield biased predictions, when $\lambda$ is too large, the biased
information from each single view can no longer be effectively eliminated
during integration, which can hurt the overall performance.




\noindent \textbf{Effect of $\mu$.} Figure \ref{fig:mu} shows the effect of
$\mu$.  Similarly, as $\mu$ increases, the performance of \ours first increases
and then decreases when $\mu$ is too large.  The reason for this phenomenon is
that: 1) when $\mu$ is too small, the three models cannot regularize each other
well, which hinders them from sharing the result with others; 2) when the $\mu$ is too large, then the output will be close to optimizing Equation \ref{eq:14}.
When one model does not perform well, it will negatively affect the other two models, which will deteriorate the performance of the overall model.

\subsection{Case Studies and Error Analysis}
\begin{figure}[t]
        \centering
            \includegraphics[width=0.48\textwidth]{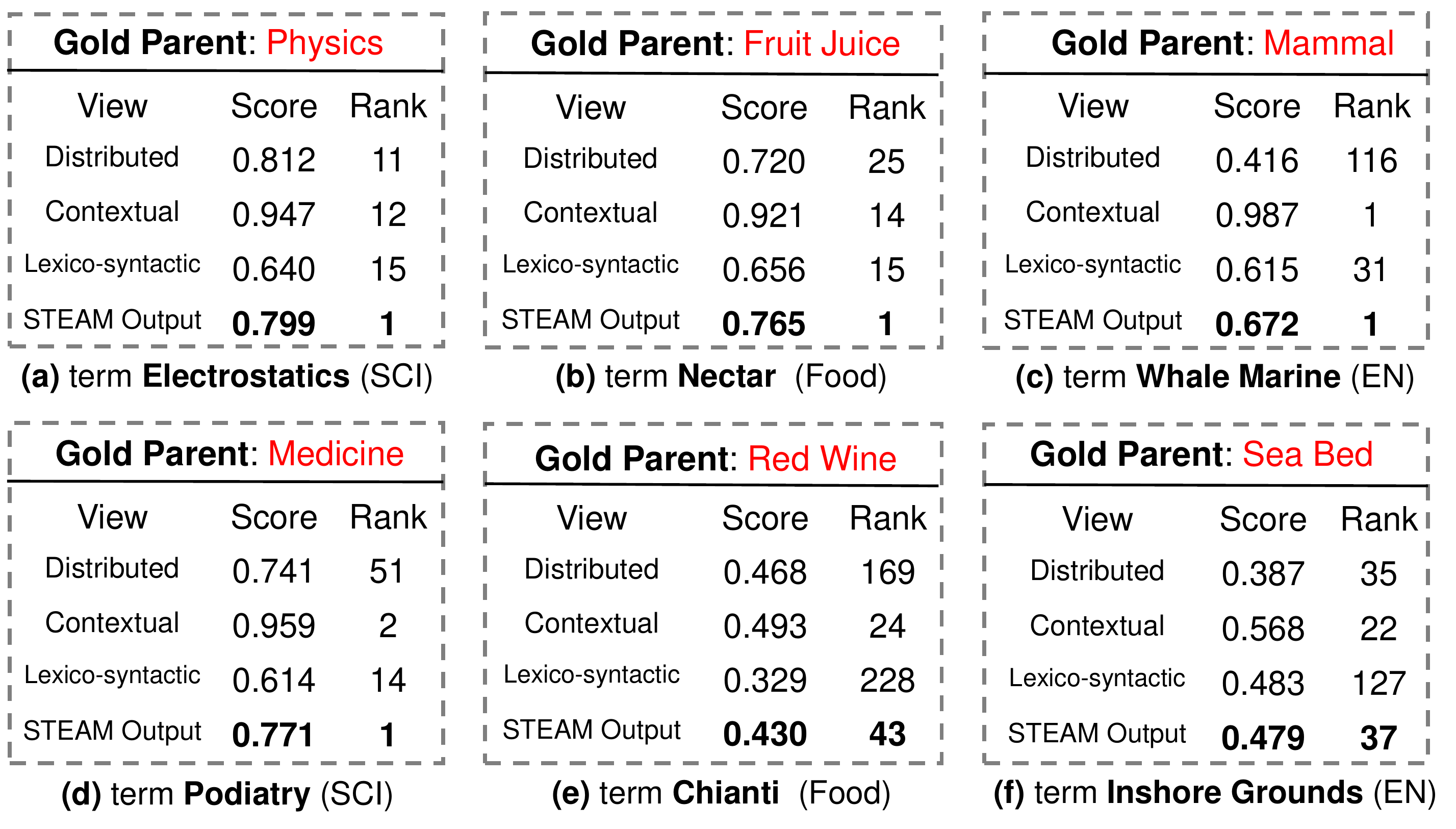}
            \vspace{-0.1in}
    \caption{Prediction result for several test terms from different datasets.}
    \vspace{-0.1in}
    \label{fig:case}
\end{figure}
Figure \ref{fig:case} shows multiple cases to illustrate the efficacy of \ours.
It reports the final prediction score of \ours for the ground-truth parent,  as
well as the prediction scores from the three base classifiers. Based on the
scores, we calculate the rank of the ground truth parent.  From
Figure~\ref{fig:case}(a), (b), we can find that there are cases when the
predictions from all the three views are inadequate, but the final prediction
can integrate the weak signals to rank the ground-truth to the top. Such cases
verify the power of multi-view co-training in \ours, which can utilize the
complementary signals from all views and improve the final performance. Besides, Figure~\ref{fig:case}(c), (d) shows two cases when the predictions
of one specific view are poor (\eg Distributed view for term \texttt{Whale
Marine}), yet \ours can rectify the mistakes by leveraging the information from
the other two views.
Figure~\ref{fig:case}(e) and (f) show two random examples on which our model
fails to provide the correct predictions. Through in-depth analysis, we found
that one common case when our model cannot perform well is that all views
cannot make accurate predictions. In such cases, the information from the three
views is insufficient to capture the hypernymy relationships between the test
term and its parent. 



\section{Related Work}



\vspace{0.5ex}
\noindent
\textbf{Taxonomy Construction.}
There have been many studies on automatic taxonomy construction. One line of
works constructs taxonomies using cluster-based methods.  They group terms into a
hierarchy based on hierarchical clustering \cite{alfarone2015unsupervised, zhang2018taxogen,shang2020nettaxo} or topic models \cite{liu2012automatic, downey2015efficient}.
These methods can work in an unsupervised way. However, they cannot be applied
to our taxonomy expansion problem, because they construct \emph{topic-level}
taxonomies where each node is a collection of topic-indicative terms instead of
single terms.  More relevant to our work are the methods developed for
constructing \emph{term-level} taxonomies.  Focused on taxonomy induction,
these methods organize hypernymy pairs into taxonomies.  Graph optimization
techniques \cite{kozareva2010semi,gupta2017taxonomy,bansal2014structured,cocos2018comparing} have
been proposed to organize the hypernymy graph into a hierarchical structure,
and~\citeauthor{mao2018end}~\cite{mao2018end} utilize reinforcement learning
to organize term pairs by optimizing a holistic tree metric over the training
taxonomies. Very recently, \citeauthor{chao2020-g2t}~\cite{chao2020-g2t}~design a transfer framework to use the knowledge from existing domains for generating taxonomy for a new domain.
However, all these methods attempt to construct taxonomies
\emph{from scratch} and cannot preserve the structure of the seed
taxonomy.

\vspace{0.5ex}
\noindent
\textbf{Hypernymy Detection.}
Hypernym detection aims at identifying hypernym-hyponym pairs,
which is essential to taxonomy construction.
Existing methods for hypernymy detection mainly fall into two
categories: \emph{pattern-based} methods and \emph{distributed} methods.
\emph{Pattern-based} methods extract hypernymy pairs via
pre-defined lexico-syntactic patterns \cite{hearst1992automatic, panchenko2016taxi,roller2018hearst}. One prominent work in this branch is the
Hearst patterns~\cite{hearst1992automatic}, which extract hypernymy pairs based
on a set of hand-crafted \textit{is-a} patterns (\eg, ``X is a Y'').
Pattern-based methods achieve good
precision, but they suffer from low recall \cite{wu2012probase} and are prone to
idiomatic expressions and parsing errors \cite{kozareva2008semantic}.
\emph{Distributed} methods detect hypernymy pairs based on the
distributed representations (\eg word embeddings \cite{mikolov2013distributed,pennington2014glove,devlin2019bert}) of terms.
For a term pair \emph{$\langle
x,y\rangle$}, their embeddings are used for learning a binary classifier to predict whether it has the hypernymy relation
\cite{baroni2012entailment,fu2014learning,chang2018distributional,shi2019discovering}.  As embeddings are directly learned from the corpora, distributed methods eliminate
the needs of designing hand-crafted patterns and have shown strong performance.
However, their performance relies on a sufficient amount of labeled term pairs,
which can be expensive to obtain.



\vspace{0.5ex} \noindent \textbf{Taxonomy Expansion.} Taxonomy expansion is less
studied than taxonomy construction. Most existing works on taxonomy expansion
aims to find new \emph{is-a} relations and insert new terms to their hypernyms.
For example, \citeauthor{aly2019every}~\cite{aly2019every} refine existing taxonomy by
adopting hyperbolic embeddings \cite{nickel2017poincare} to better capture
hierarchical lexical-semantic relationships,~\cite{vedula2018enriching,shen2018hiexpan} design various semantic patterns to determine the position to attach new concepts for expanding taxonomies, and \citeauthor{fauceglia2019automatic}~\cite{fauceglia2019automatic} use a hybrid method to take advantage of linguistic patterns, semantic web and neural
networks  for taxonomy expansion. However, the above methods only model the `parent-child' relations and fail to capture the global structure of the existing taxonomy. 
To better exploit self-supervision signals, \citeauthor{Manzoor2020expand}~\cite{Manzoor2020expand} study expanding
taxonomies by jointly learning latent representations for edge semantics and
taxonomy concepts. Recently,
\citeauthor{shen2020taxoexpan}~\cite{shen2020taxoexpan} propose position-enhanced graph neural networks to encode the relative position of terms
and improve the overall quality of taxonomy. 
However, the above two approaches only consider distributional features such as word embeddings but neglect other types of relationships among terms.
Compared with these methods, \ours is novel in two aspects. \emph{First}, it inserts new terms with mini-path-based classification instead of simple hypernym attachment, which models different layers to better preserve the holistic structure. \emph{Second}, it considers multiple
sources of features for expansion  and integrates them with a multi-view co-training procedure.




\section{Conclusion}


We proposed \ours, a self-supervised learning framework with novel
mini-path-based prediction and a multi-view co-training objective.  The
self-supervised learning nature enables our model to optimize the utilization
of the knowledge in the existing taxonomy without labeling efforts.  Compared
with the traditional node-to-node \textit{query-anchor} pairs, the adoption of
mini-paths captures more structural information thus facilitates the inference
of a query's attachment position.  The multi-view co-training objective
effectively integrates information from multiple input sources, including
PGAT-propagated word embeddings, LSTM-embedded dependency paths and
lexico-syntactic patterns.  Comprehensive experiments on three benchmarks show
that \ours consistently outperforms all baseline models by large margins, which
demonstrates its superiority for taxonomy expansion. 

\section*{Acknowledgement}
This work was in part supported by the National Science Foundation award IIS-1418511, CCF-1533768 and IIS-1838042, the National Institute of Health award 1R01MD011682-01 and R56HL138415.
\section*{Appendix--External Sources of Corpus}
\ours (our method) and several baselines require external text corpora to
model the semantic relations between concept terms.
For all the three
benchmarks, we collect the following public corpora: 1) the Wikipedia
dump\footnote{We use the 20190801 version of wikidump during our experiments.},
2) the UMBC web-based
corpus\footnote{https://ebiquity.umbc.edu/resource/html/id/351}; 3) the One
Billion Word Language Modeling
Benchmark\footnote{https://www.statmt.org/lm-benchmark/}.

We directly match the terms with the corpus with tools available online (\ie WikiExtractor\footnote{https://github.com/attardi/wikiextractor}) and only preserve the sentences that term pairs co-occur. In this way, for each dataset, we obtain a tailored corpus which preserves the co-occurrence between terms. The information for these corpora are summarized as:

\noindent $\bullet$ \textbf{Environment}: The corpus size is  824MB with 1.51M sentences.

\noindent $\bullet$ \textbf{Science}: The corpus size is  1.36GB with 2.07M sentences.

\noindent $\bullet$ \textbf{Food}: The corpus size is  2.00GB with 3.42M sentences.

\bibliographystyle{ACM-Reference-Format}
\bibliography{refer}

\end{document}